\documentclass[10pt,twocolumn,letterpaper]{article}

\usepackage[pagenumbers]{wacv} % To force page numbers, e.g. for an arXiv version

\usepackage{graphicx}
\usepackage{amsmath}
\usepackage{amssymb}
\usepackage{booktabs}

\usepackage{amsmath}
\usepackage{algorithm}          % for pseudocode
\usepackage{algpseudocode}      % for pseudocode
\usepackage{multirow}           % table
\usepackage{amssymb}            % add. symbols
\usepackage{pifont}             % add. symbols
\usepackage{graphicx}
\usepackage{forest}
\usepackage{tabularx}
\newcolumntype{C}{>{\centering\arraybackslash}X}
\usepackage{enumitem}
\definecolor{tabfirst}{rgb}{1, 0.7, 0.7} % red
\definecolor{tabsecond}{rgb}{1, 0.85, 0.7} % orange
\definecolor{tabthird}{rgb}{1, 1, 0.7} % yellow
\usepackage[accsupp]{axessibility}

\usepackage[pagebackref,breaklinks,colorlinks]{hyperref}

\usepackage[capitalize]{cleveref}
\crefname{section}{Sec.}{Secs.}
\Crefname{section}{Section}{Sections}
\Crefname{table}{Table}{Tables}
\crefname{table}{Tab.}{Tabs.}

\def\wacvPaperID{2128} % *** Enter the WACV Paper ID here
\def\confName{WACV}
\def\confYear{2025}

\begin{document}

\title{SEED4D: A Synthetic Ego--Exo Dynamic 4D Data Generator, \\  Driving Dataset and Benchmark}

\author{%
  \textbf{Marius Kästingschäfer}$^{1,2}$ 
  \quad
  \textbf{Théo Gieruc}$^{1}$
  \quad
  \textbf{Sebastian Bernhard}$^{1}$\\
  % \quad
  \textbf{Dylan Campbell}$^{3}$
  \quad
  \textbf{Eldar Insafutdinov}$^{4}$
  \quad
  \textbf{Eyvaz Najafli}$^{1,5}$
  \quad
  \textbf{Thomas Brox}$^{2}$\\
  $^1$Continental
  \quad
  $^2$University of Freiburg
  \quad
  $^3$Australian National University\\
  % \quad
  $^4$University of Oxford 
  \quad
  $^5$University of Tübingen\\
  \texttt{marius.kaestingschaefer@continental.com}
}

\twocolumn[{%
\renewcommand\twocolumn[1][]{#1}%
\maketitle
\begin{center}
    \centering
    \captionsetup{type=figure}
    \includegraphics[width=1\textwidth]{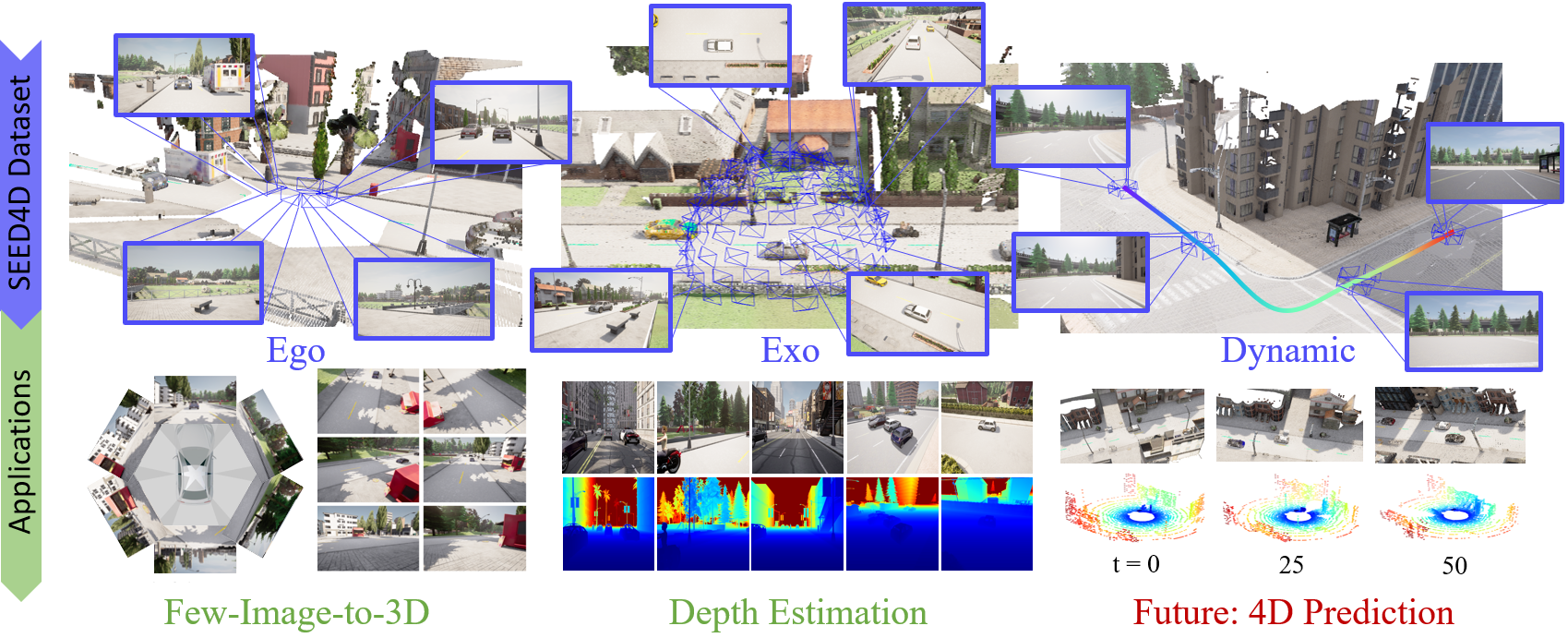}
    \captionof{figure}{The SEED4D dataset contains synthetic egocentric--exocentric dynamic 4D data and pose information (top). We benchmark existing novel view synthesis, depth, and few-image-to-3D methods and propose 4D prediction as a future open challenge (bottom).}
    \label{fig:overview_paper}
\end{center}
}]

\maketitle

%%%%%%%%% ABSTRACT
\begin{abstract}
    Models for egocentric 3D and 4D reconstruction, including few-shot interpolation and extrapolation settings, can benefit from having images from exocentric viewpoints as supervision signals. No existing dataset provides the necessary mixture of complex, dynamic, and multi-view data. To facilitate the development of 3D and 4D reconstruction methods in the autonomous driving context, we propose a Synthetic Ego--Exo Dynamic 4D (SEED4D) data generator and dataset. We present a customizable, easy-to-use data generator for spatio-temporal multi-view data creation. Our open-source data generator allows the creation of synthetic data for camera setups commonly used in the NuScenes, KITTI360, and Waymo datasets. Additionally, SEED4D encompasses two large-scale multi-view synthetic urban scene datasets. Our static (3D) dataset encompasses 212k inward- and outward-facing vehicle images from 2k scenes, while our dynamic (4D) dataset contains 16.8M images from 10k trajectories, each sampled at 100 points in time with egocentric images, exocentric images, and LiDAR data. The datasets and the data generator can be found \href{https://seed4d.github.io/}{here}.
\end{abstract}
\vspace{-0.4cm}

%%%%%%%%% BODY TEXT
\section{Introduction}\label{sec:intro}

Within robotics and especially autonomous driving, inferring the 3D environment \cite{li2023delving, xu2024survey} and making predictions about the temporal evolution of a scene \cite{liu2021survey, Mozaffari_2022} is essential for operating safely. Tasks associated with those problems such as video prediction \cite{yan2023temporally_teco, yan2021videogpt, seo2022harp, harvey2022flexible, Oprea_2022_video_prediction_review}, point cloud forecasting \cite{zhang2024learning_autopilot4D, spf2_weng2020inverting, s2net_pointcloud_forecasting_weng, khurana2023point4D}, and few-image-to-3D reconstruction \cite{yu_pixelnerf_2021, szymanowicz23splatter, gieruc20246imgto3d} are currently approached separately. 
% shortcomings of existing datasets:
Jointly performing these tasks requires a comprehensive datasets consisting of a diverse and extensive array of non-egocentric vehicle images collected in dynamic scenes. Current datasets lack such a mixture of images.
% Viewpoint inadequacy
Most autonomous driving datasets offer only egocentric vehicle viewpoints. These are insufficient to supervise reconstructed birds-eye views or third-person vehicle views. Training a few-image-to-3D or a 4D prediction model using purely outward-facing camera images might be possible, but evaluating the generalizability of the trained models requires several \textit{non-ego supervision views} from diverse viewpoints.
% temporal data scarcity/viewpoint limitations: 
While most 3D reconstruction datasets comprise a large number of viewpoints, they \cite{synthetic_NeRF_dataset, tanks_temples_knapitsch2017} lack what most autonomous driving datasets contain, namely a large amount of \textit{temporal data} from heterogeneous scenes. Some datasets are restricted to single objects \cite{chang2015shapenet, objaverse, zhou2016thingi10k, objaverseXL, Lu_2024_CVPR}, scenes with limited shape and texture complexity \cite{dmlab_zhai2020largescale, synthetic_NeRF_dataset, KTH_Action_2004}, or data with very few short camera videos \cite{mildenhall2019llff, park2021hypernerf, park2021nerfies}. However, many unbounded and diverse training scenes are required to train large-scale few-image-to-3D architectures or 4D forecast models. To the best of our knowledge, currently no large-scale egocentric--exocentric data generator or datasets for autonomous driving exists.

% why it is important / scientific reasoning for paper
We further observe a gap between temporal prediction methods and spatial reconstruction methods. Currently, there is little interaction between spatiotemporal reconstruction methods \cite{fridovich-keil_k-planes_2023, cao2023hexplane, li2022neural3D, li2021neural, du2021neural, yunus2024recent} and video prediction methods \cite{yan2023temporally_teco, yan2021videogpt, seo2022harp, harvey2022flexible, Oprea_2022_video_prediction_review}. While spatiotemporal reconstruction methods deal with recreating and encoding 4D scenes, video prediction methods predict the next 2D camera frame. Due to explicit 3D modeling, reconstruction methods offer free control over camera movements, potentially resulting in more accurate geometric representations. Especially within autonomous driving, the spatiotemporal reconstruction of large-scale scenes is being investigated \cite{zhou2024drivinggaussian, yan2024street, zhou2024hugs}. Common video prediction methods tokenize 2D image inputs and perform autoregressive predictions using transformer or transformer-diffusion-based architectures \cite{peebles2023scalable, gupta2023photorealistic, liu2024sora}. While diffusion-based 2D and 3D models produce partially geometrically consistent outputs \cite{sarkar2023shadows, huang2023vbench, li2024sora}, they generally lack camera control and spatiotemporal consistency. Existing 4D prediction methods are limited to point cloud forecasting \cite{zhang2024learning_autopilot4D, spf2_weng2020inverting, s2net_pointcloud_forecasting_weng, khurana2023point4D}. Such predictions, however, lack visual fidelity and need sensors other than simple RGB cameras. Given the recent acceleration of developments in both 3D reconstruction and video prediction methods and the apparent shortcomings of existing methods, transitioning towards 4D predictions and facilitating this development with the introduction of a new data generator and a first dataset appears scientifically justified.

% What our dataset does better
We introduce a \textbf{S}ynthetic \textbf{E}gocentric--\textbf{E}xocentric \textbf{D}ynamic \textbf{4D} data generator and dataset (SEED4D). SEED4D consists of a data generator and two datasets to address the aforementioned shortcomings. We additionally propose a few-image-to-3D reconstruction and novel view synthesis benchmarks.
% scalability and control
To streamline the development of 3D and 4D research, we propose an easy-to-use, 
customizable Ego-Exo view data generator. Our framework provides a plug-and-play solution for generating novel spatial and temporal driving data, making it easy for practitioners and researchers to create personalized datasets quickly. Our data generator tackles data scarcity by enabling flexible, fine-grained viewpoint control and multi-camera data collection over an extended period. 
The provided viewpoints can be collected from any vehicle or other point in the scene. Our data generator this way enables the generation of datasets for 3D or 4D prediction tasks.
% dense annotations
Beyond volume, due to the CARLA Simulator \cite{Carla_Dosovitskiy17}, our data generator and the data generator provide reliable ground truth annotations of depth, optical flow, instance, and semantic segmentation together with 3D LiDAR point clouds. We output all pose information in a NeRFStudio-suitable \cite{nerfstudio_Tancik_2023} format, simplifying data usage. Using synthetic data is sensible, since it involves fewer ethical concerns, enables reproducibility, and is easily scalable. Collecting real-world ego--exo data similar to ours would be very costly. Further, domain transfer methods can be used to reduce the gap in appearance \cite{huang2018multimodal, zhu2020unpaired, keser2021content, dmlab_zhai2020largescale, wang2023stylediffusion}, and zero-shot transfer after the geometric learning task has shown promising results \cite{gieruc20246imgto3d}.
% few-image-to-3D
We provide two example datasets to highlight the flexibility of our data generator. Our datasets comprise high-resolution \textit{egocentric and exocentric} (i.e., non-egocentric) vehicle camera data, ideally suited to train few-image-to-3D models when using non-ego target views. This task is not possible with existing autonomous driving datasets.
% temporal multi-view data
SEED4D also contains complex multi-view ego--exo images from dynamic urban environments (traffic, pedestrians, weather) and thus provides the \textit{spatiotemporal richness} lacking in existing reconstruction datasets. By capturing both ego--exo multi-viewpoint and multi-timestep (dynamic) data, our resulting dataset can aid models in improving temporal consistency.

\begin{table*}[!ht]
    \centering
    \caption{Comparison of different autonomous driving (AD) datasets. The last two rows in the table showcase the static and the dynamic dataset obtained using our data generator. EgoV denotes ego views, and 3rdPV stands for 3rd person views.}
    \label{AD_dataset_comparison}
        \begin{tabular}{ l c c c c c c c c c c}
        \toprule
              \textbf{Datasets}    	                   & \textbf{\# Seq.} & \textbf{Length (s)}          & \textbf{EgoV}     & \textbf{3rdPV}          & \textbf{Depth}          & \textbf{LiDAR}         & \textbf{3D Bbox}  & \textbf{Type}\\
        \midrule 
              Cityscapes \cite{cordts2016cityscapes}       & 46          & 1.8       & 25k        &  $\times$      & $\times$     & $\checkmark$  & $\times$ & Real-World \\
              KITTI \cite{KITTI_geiger2012, Geiger2013IJRR}& $\sim$ 330  & $\sim$ 65 & $\sim$ 61k &  $\times$      & $\times$     & $\checkmark$  & $\checkmark$   & Real-World  \\
              KITTI360 \cite{KITTI360_Liao2022}            & n/a         &  n/a      & 300k       &  $\times$      & $\times$     & $\checkmark$  & $\checkmark$ & Real-World\\
              NuScenes \cite{caesar2020nuscenes}           & 1k          & 20        & 1.4M       &  $\times$      & $\times$            & $\checkmark$  & $\checkmark$ & Real-World\\
              ARGOverse \cite{chang2019argoverse}          & 1k          & 15        & 2.7M       &  $\times$      & $\times$     & $\checkmark$  & $\checkmark$ & Real-World\\
              Waymo Open \cite{waymo_sun2020scalability}   & 1k          & 20        & 1M         &  $\times$      & $\times$              & $\checkmark$  & $\checkmark$ & Real-World\\
              BDD100K \cite{yu2020bdd100k}                 & 100k        &  n/a       & 100M      &  $\times$      & $\times$     & $\times$      & $\times$     & Real-World \\
              SEED4D (Ours)                                 & 2k         &  n/a       & 12k       & \textbf{200k}  & $\pmb{\checkmark}$ &$\pmb{\checkmark}$  & $\pmb{\checkmark}$ & Synthetic\\
              SEED4D (Ours)                                 & 10k        &  10        & 6.3M      & \textbf{10.5M} & $\pmb{\checkmark}$ & $\pmb{\checkmark}$  & $\pmb{\checkmark}$ & Synthetic\\
        \bottomrule
        \end{tabular}
\end{table*}

% contributions
We summarize the four key contributions of this paper as follows:

\begin{enumerate}[leftmargin=*,nosep]
    \item \textbf{Data Generator.} We present a customizable data generator based on the CARLA autonomous driving simulator outputting NeRFStudio suitable intrinsic and extrinsic camera poses. We provide several pre-specified camera setups to generate datasets similar to NuScenes, KITTI360, and Waymo, which consist of ego and surround vehicle sensor suits. 
    \item \textbf{Static Dataset.} We provide a pre-generated dataset of 2k unbounded outdoor driving scenes with 100 inward-facing exocentric images and more than six out-of-vehicle images for each, resulting in a total of 212k images.
    \item \textbf{Dynamic Dataset.} This spatiotemporal dataset consists of 10.5k individual trajectories, each of 100 timesteps, resulting in a length of 10 seconds per trajectory. This dataset encompasses images of multi-vehicle ego and non-ego vehicle views and detailed pose information.
    \item \textbf{Benchmarks.} We perform an evaluation of existing methods on the proposed datasets. We choose a few-image-to-3D tasks and a novel view synthesis task. For several methods, we also measure the quality of the scene reconstruction. The benchmarks offer challenging tasks in an unbounded autonomous driving environment.
\end{enumerate}

The data generator and the datasets are released openly to support the development of few-image out-of-vehicle reconstruction methods and 3D-aided temporal prediction models. An overview of SEED4D, along with with links to the code and the dataset, is available on our \href{https://seed4d.github.io/}{project page}.

\begin{figure*}
    \centering
    \includegraphics[width=1\linewidth]{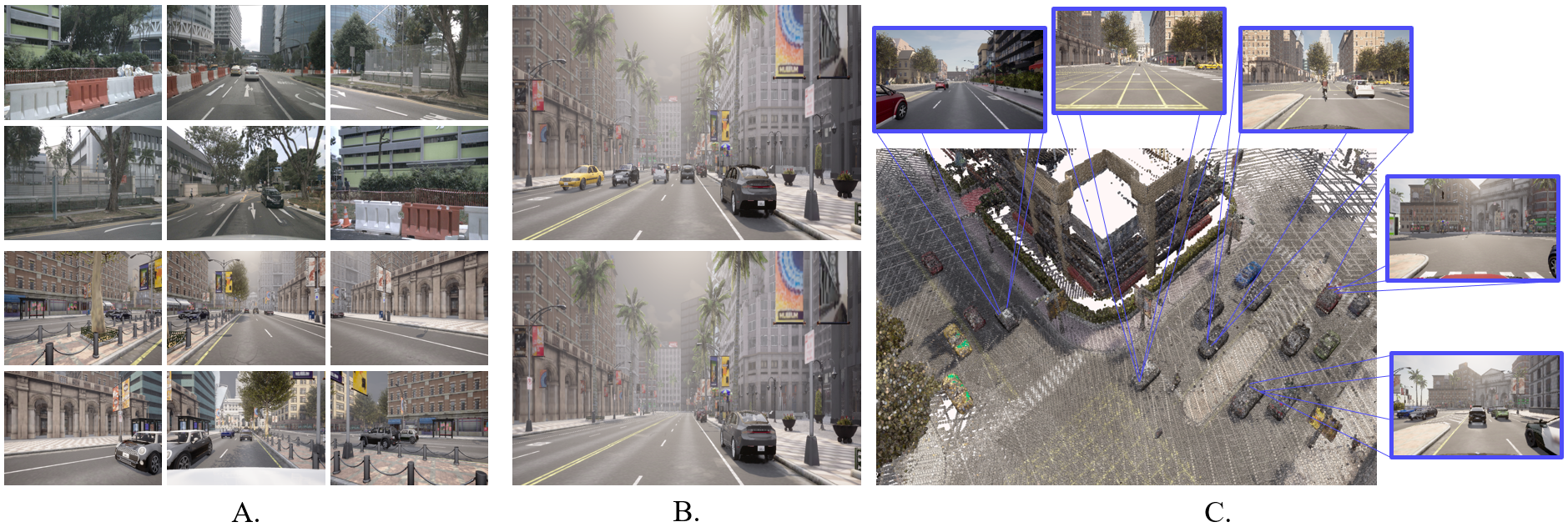}
    \caption{\textbf{Data Generator Capabilities.} \textbf{A.} Showing example nuScenes images \cite{caesar2020nuscenes} and images generated by our data generator with similar intrinsic and extrinsic pose information. \textbf{B.} Removing dynamic vehicles from the scene. The parked vehicles remain. \textbf{C.} Generating egocentric views for all vehicles in a scene.}
    \label{fig:generator_capabilities}
\end{figure*}

\section{Related Work}\label{sec:background}

The spatio- and temporal data our data generator can produce together with the datasets we introduce are at the intersection of 3D, 4D, and autonomous driving. Often 3D datasets consist of several camera views per timestep or consist of static scenes filmed over time with a moving camera, while 4D datasets are characterized by the underlying scene being dynamic and evolving. These datasets encompass scenarios where one or multiple cameras, themselves static or in motion, observe the scene.

\textbf{3D Datasets.} In recent years, the number of available datasets for 3D reconstruction has increased. Most of these are also suitable for few-image-to-3D tasks. 
% Synthetic Multi-View Single Object Datasets:
Datasets introduced for training novel view synthesis methods often focus on forward-facing scenes or inward-facing camera setups \cite{synthetic_NeRF_dataset, mildenhall2019_LLFL, jensen_large_2014, yao2020blendedmvs}, where the camera is moved in proximity to an object or a scene. Several datasets focus on single objects in a bounded environment or with a white background \cite{chang2015shapenet, objaverse, zhou2016thingi10k, objaverseXL}. Many showcase a variety of household objects \cite{AsaadiMK19_BigBird, Calli_2015_YCB, sun2018pix3d} and common objects \cite{reizenstein2021common}. Both synthetic \cite{tremblay2022rtmv, tanks_temples_knapitsch2017} and real-world \cite{downs_google_2022} objects and images are widely used.
% Large-Scale Scenes
Another category of datasets focuses on indoor \cite{dai2017scannet, straub2019replica, Yu2023PhotoconsistentNVS}, unbounded outdoor \cite{aanaes2016large, ling2023dl3dv10k} or mixed scenes \cite{ling2023dl3dv10k}. The NeRDS360 dataset \cite{irshad_neo_2023} resembles our data in some aspects. NeRDS360 also provides surround vehicle supervision images of a driving scene. However, this dataset unlike ours does not provide any ego-vehicle views, only covers 75 scenes, and does not include temporal scenes or LiDAR data.
% egocentric data
Furthermore, several indoor and outdoor datasets exist, either containing only first-person observations \cite{yan2023temporally_teco, wang2023embodiedscan, li2024egogen}, third-person views \cite{lv2024aria} or ego--exo person views \cite{grauman2023egoexo4d}. While those are partially also used for video prediction tasks \cite{yan2023temporally_teco}, none of them except for NeRDS360 is autonomous driving specific.

\textbf{4D Datasets.} Due to success in modeling 3D scenes, many models have moved towards including the temporal domain.
% multiple + single cameras
Commonly used multi-view real-world datasets often have more than ten cameras per timestep \cite{broxton2020immersive, li2022neural3D} whereas other data collections focus on hand-held cellphone cameras \cite{mildenhall2019llff, park2021hypernerf, park2021nerfies}.
% humans
Additional supervising signals such as depth \cite{božič2020deepdeform} or 4D mesh information \cite{dfaust_2017} are occasionally provided. Many of the existing dynamic data collections are human \cite{Habermann_2021_DynaCap, zheng2022structured, dfaust_2017, bhatnagar22behave} or animal-centered \cite{sinha2022common}, mainly focusing on human poses \cite{3DPW_Marcard_2018}.
% ego views
Such human-related data also exists from an ego perspective \cite{Damen2018EPICKITCHENS, EPICFields2023}. 
% synthetic data
Commonly synthetic data is also used for 4D datasets \cite{nanbo2020mulmon, hu2021sailvos, li20214dcomplete}.
% video prediction
For video prediction tasks, datasets range from small-scale ones \cite{KTH_Action_2004, srivastava2015unsupervised} to large-scale \cite{carreira2018short, yan2023temporally_teco} and very large \cite{abuelhaija2016youtube8m, webvid_Bain21}. They are, however, largely unstructured, offer no additional pose or camera information, and are not automotive-specific.

\textbf{Autonomous Driving Datasets.} Numerous real-world autonomous driving datasets exist. Most of them provide a car-centric first-person view and come with several additional sensors. 
Table \ref{AD_dataset_comparison} summarizes the mentioned autonomous driving datasets, highlighting their viewpoint and sensor coverage. A unique feature of our datasets is the ego--exo views, a combination that does not exist in common real-world datasets.
% Argoverse
Existing autonomous driving datasets such as Argoverse often include images from multiple RGB cameras, LiDAR information, and 3D bounding boxes \cite{chang2019argoverse}. 
% Cityscapes
An exception is the Cityscapes \cite{cordts2016cityscapes} dataset that contains only front-facing vehicle views similar to the massive BDD100K dataset \cite{yu2020bdd100k} and the YouTube-scraped OpenDV-YouTube \cite{yang2024generalized} data collection.
% Kitti, Semantic Kitti, and KITTI360
Works such as KITTI provide additional measurements such as GPS information \cite{Geiger2013IJRR}, 3D semantic occupancy values \cite{behley2019iccv, behley2021ijrr, geiger2012cvpr} or rich sensory 3D annotations \cite{Liao2022PAMI}. 
% NuScenes
Datasets such as NuScenes \cite{caesar2020nuscenes} or BlockNeRF \cite{blocknerf_dataset} provide a 360-degree surround vehicle view and broad scene coverage.
% Waymo
The sequence length obtained from different autonomous driving datasets differs widely. Some of the longest sequences are for example the ones within the Waymo Open dataset which span 20 seconds, each to a large degree annotated and with calibrated LiDAR information.
% Occ3D
A purely 3D semantic occupancy-based autonomous driving dataset is Occ3D \cite{tian2023occ3d}, which only aimed toward semantic occupancy prediction methods. 
% usp seed4d
It is important to note that none of the mentioned autonomous driving datasets includes non-ego views or other privileged 3D information about occluded regions. We are aware of only one paper attempting to reconstruct the surroundings using multiple ego-views from Argoverse 2 data \cite{fischer2024multilevel}. However, this sub-dataset only comprises two scenes and does not include multiple viewpoints at the same timestep, and different lighting and weather conditions make evaluating dynamics prediction models in this setting difficult. \

% synthetic AD datasets
Synthetically generated autonomous driving datasets supplement real-world data, offering controlled environments and diverse scenarios. They either consist of static images \cite{Ros_2016_SYNTHIA, wrenninge2018synscapes} or dynamic scenes. Such datasets are created using a wide variety of simulators \cite{shah2017airsim, zhou2024garchingsim, richter2016playing} such as CARLA \cite{Carla_Dosovitskiy17}, a widely utilized open-source platform offering realistic kinematic parameters and comprehensive documentation. 
Most synthetic data is richly annotated \cite{richter2016playing, VIPER_Richter_2017} and purpose-built for tasks such as vehicle tracking as Synthehicle \cite{herzog2022synthehicle}. Similar to the NERDS360 dataset \cite{irshad_neo_2023}, Synthicle consists of a large number of exo views but does not contain any ego views. Other datasets are designed for evaluating cross-lane novel view synthesis \cite{li2024xldcrosslanedatasetbenchmarking}, adversarial robustness \cite{nesti2022carlagear} or replicating KITTI virtually \cite{gaidon2016virtual_KITTI1, cabon2020virtualKITTI2, deschaud2021kitticarla} but without providing any non-ego vehicle views.

\begin{figure*}[!ht]
    \centering
    \includegraphics[width=\linewidth]{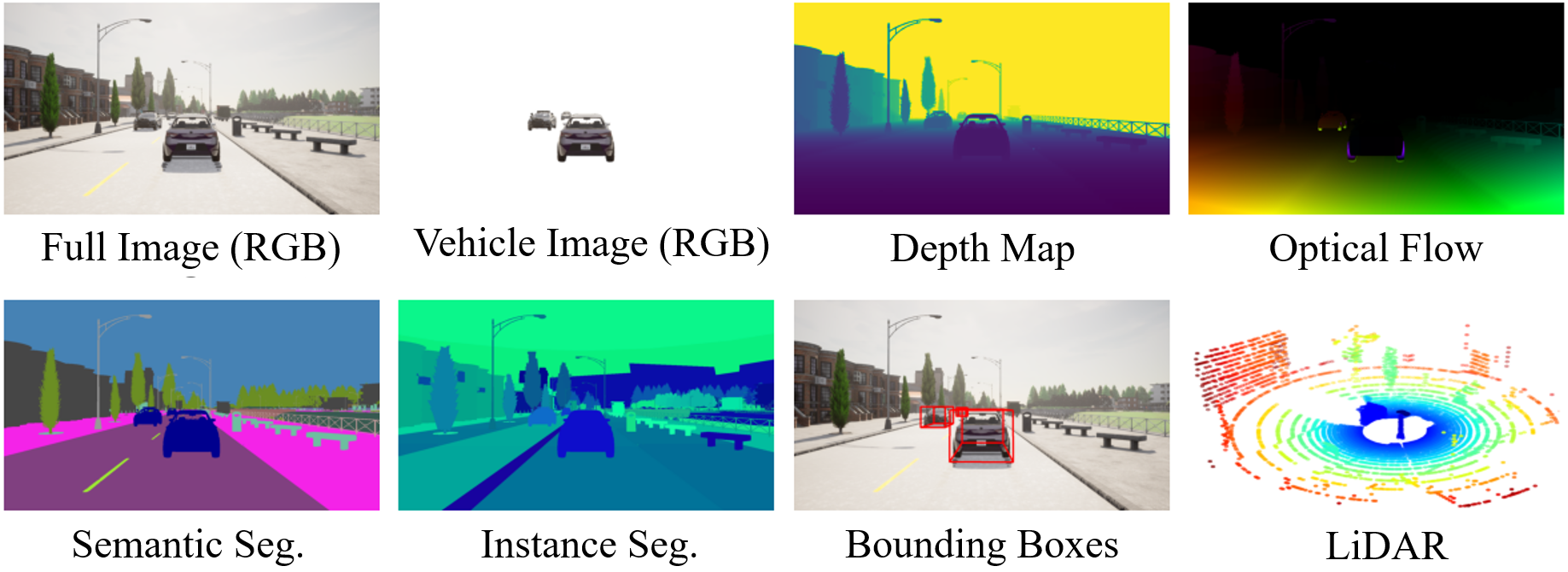}
    \caption{Overview of sensor data contained within the SEED4D datasets.}
    \label{fig:sensors}
\end{figure*}

\begin{figure}[ht!]
    \centering
    \includegraphics[width=\linewidth]{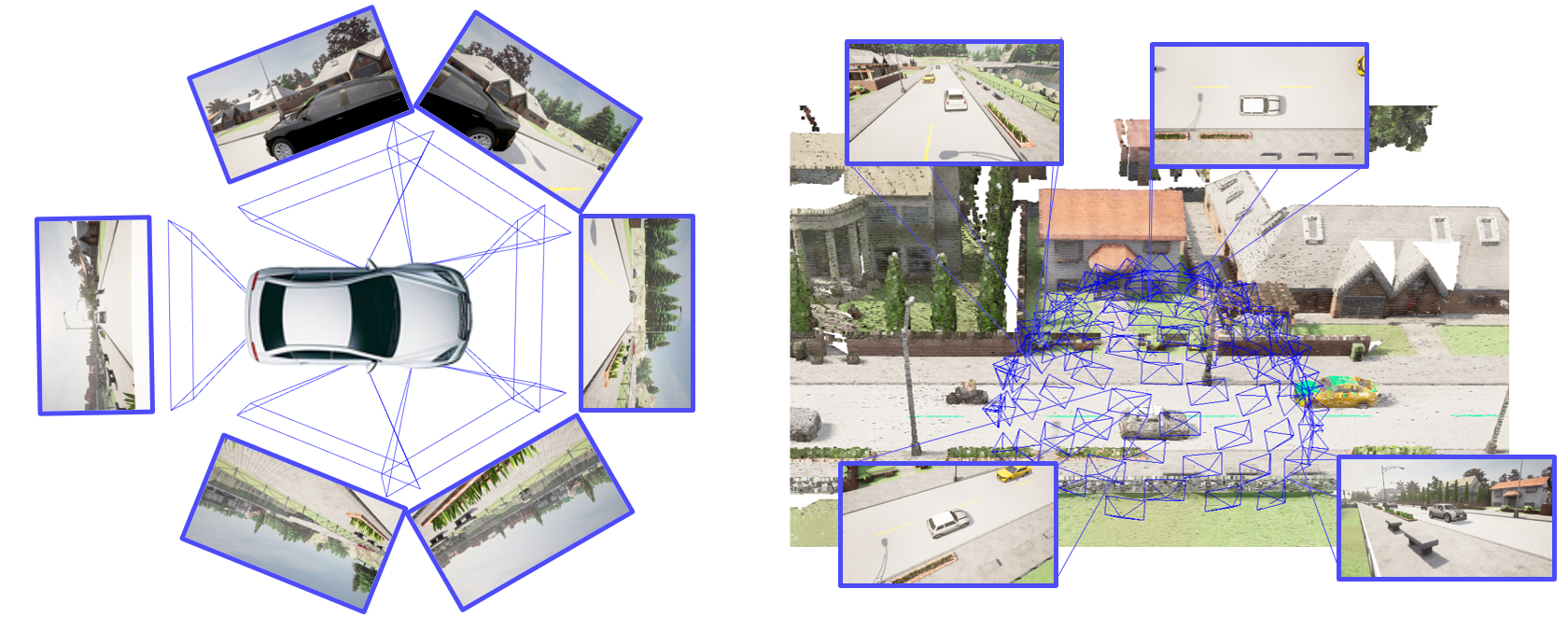}
    \caption{Egocentric and exocentric sensor configuration. The six egocentric views have a FoV of $90^{\circ}$ (the seventh $110^\circ$ rear camera is not shown here). The exocentric views also have a FoV of $90^{\circ}$ and are positioned on a half-sphere oriented towards the vehicle at the sphere center.}
    \label{fig:sensor-setup}
\end{figure}

\section{SEED4D}\label{sec:methods}

\subsection{SEED4D Data Generator} \label{sec:generator}

The data generator provides an easy-to-use 3D and 4D data creation tool. With our data generator, one can easily define parameters such as the town, the vehicle's initial position, the weather, the number of traffic participants, the number and kinds of sensors, and their position (both ego and exocentric). The resulting data, such as images, point clouds, 3D bounding boxes, and sensor extrinsic and intrinsic values, are stored conveniently. Our primary contribution in this regard is that the open-source generator is an easy-to-use tool that makes obtaining synthetic autonomous driving scenes from numerous viewpoints straightforward. The data generator can collect data from ego and non-ego vehicles in dynamic or static scenes. A number of use cases are visualized in Figure \ref{fig:generator_capabilities}.

To generate exo vehicle views, we use a half-sphere surrounding the vehicle. The procedure we use to generate the exocentric views is based on the spherical Fibonacci lattice generation, for example, described in \cite{fibonacci_grids, spherical_fibonacci_mapping}. The procedure distributes points evenly on the surface of a sphere \cite{spherical_fibonacci_latice}. The algorithms are presented in detail in the Appendix. The data generator makes use of the open-source autonomous driving simulator CARLA \cite{Carla_Dosovitskiy17} in the backend.

If necessary, the resulting sphere can be scaled using the radius parameter, shifted towards the ego-positions origin, and offset in the z-direction. 
Other exo-vehicle camera formations, such as random cameras, infrastructure-based camera setups, or pedestrian views, can be added.
For the ego-views, we provide the camera files, containing rotation, translation, and intrinsic values for the following datasets: NuScenes \cite{caesar2020nuscenes}, KITTI360 \cite{KITTI360_Liao2022}, Waymo open dataset \cite{waymo_sun2020scalability}, Argoverse \cite{chang2019argoverse} and InterFuser \cite{Interfuser_shao2022safetyenhanced}. The camera setups were obtained by directly checking the camera poses or taken from the provided descriptions. All camera intrinsic and extrinsic camera poses are outputted in a NeRFStudio-suitable \cite{nerfstudio_Tancik_2023} format whereby the OpenGL/Blender coordinate convention for cameras is used. Here $-Z$ is the look-at direction, $+X$ is right, and $+Y$ is up. The saved transform files contain information about focal length, principal point, height, width, and radial distortion. 

For later training, we also provide several accessible post-processing options, such as normalizing and centering the camera coordinate for a single timestep or across multiple timesteps, splitting the images into training, evaluation, and test data, and obtaining images of vehicle objects only. To further simplify data generation, we provide a Docker image with a pre-running CARLA instance. \newline
\vspace{-0.1cm}

\paragraph{Dataset Generation.} Both datasets contained in this paper are generated using our data generator. During the data generation, we disregarded large vehicles since the sensory setup of the cameras did not fit those vehicles, and we wanted to collect viewpoints from all vehicles in the scene for the dynamic dataset. We set the number of pedestrians per scene to 20 and the weather of each scene to 'ClearNoon'. For the dynamic dataset we introduce a small random offset between one and three seconds at the beginning of the data recording such that vehicles are already moving when being recorded. The static data was synthesized using 6 A5000 GPUs with 24GB across multiple compute nodes and the dynamic data on 8 Tesla T4 GPUs with 24GB. Taking 132 hours for the static dataset and 390 hours for the dynamic dataset. 

\begin{figure*}[!t]
    \centering
    \includegraphics[width=\linewidth]{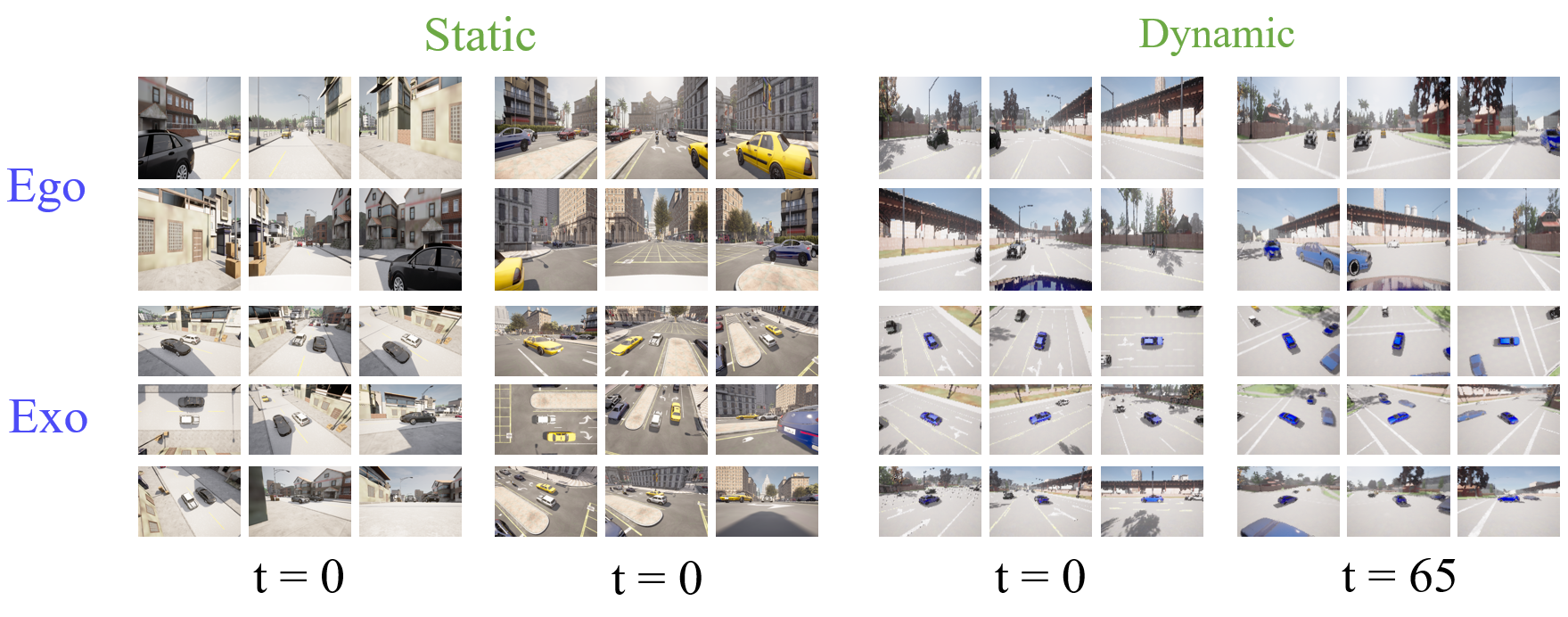}
    \caption{The left images show two scenes from the static dataset, the right images show two time points from the dynamic dataset. The images with a resolution of 16:9 were resized to fit the figure.}
    \label{fig:overview_datasets}
\end{figure*}

\subsection{SEED4D Datasets}\label{sec:dataset}

We provide two datasets: one tailored for few-image-to-3D tasks and another designed for temporal dynamics prediction tasks, both generated using our data generator. The two datasets showcase the capabilities of our data generator and also provide meaningful contributions to the community.

Each dataset $\mathcal{C}$ includes data from eight towns each with a varying number of scenes. Each scene contains $T \times N$ tuples of RGB images $\mathcal{I}_i \in \mathbb{R}^{H \times W \times 3}$ per vehicle. Here, $T$ is the number of timesteps, $N$ is the number of RGB images in the scene per timestep per vehicle and $i$ is the index of a vehicle. For each image $\mathcal{I}_i$, our dataset provides the associated extrinsic $\mathbf{E}_i=[\mathbf{R} \mid \mathbf{t}] \in \mathbb{R}^{3 \times 4}$  and intrinsic camera matrices $\mathbf{K}_i \in \mathbb{R}^{3 \times 3}$. We provide additional pixel-aligned sensor information per RGB image $\mathcal{I}_i$. Such as depth maps $\mathcal{D}_i \in \mathbb{R}^{H \times W \times 1}$, semantic segmentation masks $\mathcal{S}_{\text{sem}_{i}} \in \mathbb{R}^{H \times W \times 1}$ and instance segmentation masks $\mathcal{S}_{\text{ins}_{i}} \in \mathbb{R}^{H \times W \times 1}$. We also provide optical flow values, images containing only the vehicles, and 3D bounding boxes for the vehicles. Additionally, our datasets contain LiDAR data, which we denote as point cloud $\mathcal{P}_{i}$ consisting of $N_{P_{i}}$ points $p_{i}$. Points $p_{i}$ are tuples $(x_i, y_i, z_i, w_i)$ consisting of world coordinates $x_i, y_i, z_i$ and the intensity $w_i$ of the point. Figure \ref{fig:sensors} shows examples of the different sensors. 

In both datasets, we follow the NuScenes \cite{caesar2020nuscenes} camera configuration for the egocentric views. However, instead of using a field-of-view (FoV) of $70^{\circ}$, we use an FoV of $90^{\circ}$, which can be transformed to the NuScenes FoV ($70^{\circ}$) if required.
For the back camera, we provide a FoV of $90^{\circ}$ and $110^{\circ}$.
In this way, models can be trained under the original NuScenes settings ($70^{\circ}$ FoV cameras, $110^{\circ}$ FoV back camera) or uniform image settings ($90^{\circ}$ FoV all cameras). \\
The non-ego views are positioned around the vehicle along a half-sphere oriented towards the center where the ego vehicle is located. The cameras maintain the same absolute distance to the center vehicle. The exocentric views have a  FoV of $90^{\circ}$. The ego--exo sensor setup is visualized in Figure \ref{fig:sensor-setup}.

Alongside the sensory measurements, we provide 3D bounding boxes of all vehicles in the scene, a list of all vehicle types in the environment, a BEV gif for each vehicle collecting sensory information, and the CARLA world time elapsed. Additionally, we provide the config file with which the data are reproducible using the data generator.
\vspace{-0.2cm}

% static dataset
\paragraph{Static Ego--Exo Dataset.} We introduce a novel dataset for few-view image reconstruction tasks in an autonomous driving setting. Our dataset contains $2002$ single-timestep complex outdoor driving scenes, each offering six plus one outward-facing vehicle images and 100 images from exocentric viewpoints on a bounding sphere for supervision. Only a single vehicle in the scene is equipped with this setup. We define ego views $H_\text{ego} \times W_\text{ego}$ to be $928 \times 1600$ and the surround vehicle exo views $H_\text{exo} \times W_\text{exo}$ to be $600 \times 800$. The number of timesteps $T$ is defined to be one, and the number of image sensors $N$ is six with a $90^{\circ}$ FoV and one with a $110^{\circ}$ FoV for the ego views and 100 with a $90^{\circ}$ FoV for the exo views.
Six plus one since we save the back camera both with FOV $90^{\circ}$ and FOV $110^{\circ}$.
Other than RGB images the dataset provides $\mathcal{I}_i$ depth maps $\mathcal{D}_i$, semantic and instance segmentations $\mathcal{S}_{\text{sem}_{i}}$ and $\mathcal{S}_{\text{ins}_{i}}$. Each scene is recorded with and without the ego vehicle such that all sensors are collected twice. Across eight towns, this in 212K individual RGB images, each with its associated intrinsic camera matrix and pose. The dataset represents various driving scenes, vehicle types, pedestrians, and lighting conditions. The generated data come from Towns 1 to 7 and 10HD, resulting in 2002 unique scenes. Towns 1, 3--7, and 10HD are used for training and we left all 100 scenes from Town 2 for testing. We set the number of pedestrians and the number of non-ego vehicles to 20 each. Overall, due to the small overlap of the outward-looking ego views and the high image quality, this dataset offers a challenging task for single-shot, few-view 3D reconstruction methods.
\vspace{-0.25cm}

% dynamic dataset
\paragraph{Dynamic Ego--Exo Dataset.} Our temporal dataset consists of 10.5K driving trajectories well-suited for 4D forecasting, 4D reconstruction, or video prediction tasks. Each trajectory is 100 steps long, corresponding to a driving length of 10 seconds. The 10.5k trajectories come from a total of $498$ scenes across all towns. In each scene, the number of vehicles is set to $21$, all equipped with six plus one outward-facing vehicle camera and ten inward-facing surround vehicle exocentric images. The ego views $H_\text{ego} \times W_\text{ego}$ have size $128 \times 256$ and the exo views $H_\text{exo} \times W_\text{exo}$ are set to $98 \times 128$. Compared to the static dataset, we chose smaller image resolutions and fewer views for the non-ego views. With this image size, we tried to balance storage considerations, the amount of detail, and alignment with GPU architectures. While ego and exo cameras are non-static in the global coordinate system, their relative position to one another and the ego vehicle stay constant. We organize the dataset around a subset of all starting ego-positions, whereby only one vehicle is located directly at the specific position, and the other vehicles occupy close-by locations. To avoid all sequences starting with vehicles that are starting with a speed of zero, we introduce a small randomly sampled time offset up to three seconds. 

The static and the dynamic ego--exo view datasets are visualized in Figure \ref{fig:overview_datasets}. They differ mainly in image resolution and trajectory length and have complementary strengths. The static ego--exo dataset contains 12k egocentric views and 200k exocentric views. The dynamic ego--exo dataset contains 6.3M egocentric views and 10.5M exocentric views. More dataset details and visualizations are provided in the Appendix.

\section{Benchmarks}\label{sec:benchmarks}

Our datasets enables the comparison of existing algorithms under similar challenging conditions. We use the static dataset to compare novel view synthesis algorithms, monocular depth estimation models, and few-image-to-3D methods. We intended to compare image-based 4D prediction methods; however, we were unable to identify suitable algorithms for this purpose. Few-image-to-3D methods and novel view synthesis methods are benchmarked using established metrics such as PSNR, LPIPS \cite{zhang2018unreasonable}, and SSIM on the validation set. For the monocular metric depth estimation and the few-image-to-3D reconstruction task, we also compute the depth the root-mean-squared error (RMSE), with depth values clipped to a range of 0 to 60 meters.

\paragraph{Benchmarked Methods.}  We here briefly describe all benchmarked methods and describe their training in more detail in the Appendix. \textit{K-Planes}~\cite{fridovich-keil_k-planes_2023} factorizes a 3D scene into multiscale planes. Plane features are learned using differentiable volume rendering, sampled using multiscale bilinear interpolation, and rendered using a small MLP. We use the hybrid version of the model and the GitHub users Giodiro's reimplementation of the model available within NeRFstudio. 
\textit{NeRFacto}~\cite{nerfstudio_Tancik_2023} is a combination of several published methods. The method is optimized to work particularly well for real data captures. The following techniques are combined in this method: camera pose refinement, per-image appearance embedding learning, proposal sampling, scene contraction, and hash encoding. We use the model as part of NeRFStudio.
\textit{SplatFacto}~\cite{kerbl20233d} is a re-implementation of the original 3D Gaussian Splatting paper~\cite{kerbl20233d} within NeRFStudio. The method explicitly stores a collection of 3D volumetric Gaussians to parameterize the scene. During rendering, the 3D Gaussions are 'splatted' to obtain per-pixel colors. We use the model as part of NeRFStudio.
\textit{PixelNeRF}~\cite{yu_pixelnerf_2021} is a sparse novel view synthesis method. PixelNeRF weakens some of the shortcomings of the original NeRF paper by leveraging projected image features and training across multiple scenes. We use the re-implementation introduced in the code of Neo360 \cite{irshad_neo_2023}.
\textit{SplatterImage}~\cite{szymanowicz23splatter} is designed for inferring 3D Gaussian Splatting primitives from conditioning images in a pixel-aligned fashion. U-net style image-to-primitive mapping network supported by a cross-attention mechanism maps the input RGB images to a 'Splatter Image' containing opacity, position, shape, and color information. We use the repository released by the authors.
\textit{6Img-to-3D}~\cite{gieruc20246imgto3d} is a few-image-to-3D method specifically designed for ego--exo usecases. The method uses cross- and self-attention mechanisms during learning, projected image features during rendering, and a triplane representation as a scene representation. We use the official code release.
\textit{ZoeDepth}~\cite{bhat2023zoedepth} is a zero-shot metric depth estimation technique. The method is trained on multiple datasets, among them the autonomous driving dataset KITTI. The method builds on the MiDaS depth estimation framework. We use the publicly available code.
\textit{Metric3D}~\cite{hu2024metric3d} is a metric 3D reconstruction method using a canonical camera space transformation method. The method can perform both zero-shot metric depth and surface normal estimation from a single image. The publicly available code is used.

\paragraph{Multi-view Novel View Synthesis.} We evaluate how well existing methods can reconstruct the scene given many of the exocentric views. We divide the 100 exo views into training and test data using an 80/20 split. We evaluate the following methods contained in NeRFStudio for this task: K-Planes~\cite{fridovich-keil_k-planes_2023}, SplatFacto~\cite{nerfstudio_Tancik_2023} a reimplementation of 3D Gaussian Splatting~\cite{kerbl20233d}, and NeRFacto~\cite{nerfstudio_Tancik_2023}. The results are presented in Table \ref{tab:results_multi_img_to_3D}.

\begin{table}[!h]
    \centering
    \caption{\textbf{Multi-view Novel View Synthesis Comparison.} }
    \vspace{-0.2cm}
    \label{tab:results_multi_img_to_3D}
    \setlength{\tabcolsep}{2pt}
    \begin{tabular}{l c c c }
        \toprule
          Methods & \textbf{PSNR} $\uparrow$ & \textbf{SSIM}$\uparrow$ & \textbf{LPIPS}$\downarrow$  \\
         \midrule
          SplatFacto \cite{kerbl20233d} & \cellcolor{tabthird}24.458  &  \cellcolor{tabsecond}0.806 &  \cellcolor{tabfirst}0.210  \\
          NeRFacto \cite{nerfstudio_Tancik_2023}  & \cellcolor{tabsecond}24.936  &  \cellcolor{tabthird}0.804 &  \cellcolor{tabsecond}0.227 \\
          K-Planes \cite{fridovich-keil_k-planes_2023} &  \cellcolor{tabfirst}25.744 & \cellcolor{tabfirst}0.816  &  \cellcolor{tabthird}0.239 \\
        \bottomrule
    \end{tabular}
\end{table}

\paragraph{Monocular Metric Depth Estimation.}
Since our dataset contains ground-truth depth maps, we evaluated two recent monocular metric depth estimation methods, without fine-tuning them on our dataset.
We test the performance of Metric3D~\cite{hu2024metric3d} and ZoeDepth~\cite{bhat2023zoedepth} on the test set, namely the exocentric views of Town02. The methods we tested were used without further fine-tuning on our data. We compute the root-mean-square error of the predicted depth in meters, results are shown Table in \ref{tab:monocular_depth_results}.

\begin{table}[!h]
    \centering
    \caption{\textbf{Monocular Depth Estimation.} }
    \label{tab:monocular_depth_results}
    \begin{tabular}{l c}
        \toprule
        Methods & \textbf{DRMSE}$\downarrow$ \\
        \midrule
        ZoeDepth~\cite{bhat2023zoedepth} &  \cellcolor{tabsecond}12.352 \\
         Metric3D~\cite{hu2024metric3d} &  \cellcolor{tabfirst}7.668 \\
        \bottomrule
    \end{tabular}
\end{table}
\vspace{-0.5cm}

\paragraph{Single-shot Few-Image Scene Reconstruction.} For performing few-image-to-3D reconstruction, we deviate from many of the existing comparisons by targeting an automotive use-case and, hence evaluated the performance of methods on egocentric outward-facing views while supervising resulting novel views with  360° exocentric views. On the benchmark, we evaluate some of the previously mentioned multi-view synthesis methods and additionally the few-shot method PixelNeRF~\cite{yu_pixelnerf_2021}, SplatterImage~\cite{szymanowicz23splatter} and 6Img-to-3D~\cite{gieruc20246imgto3d}. The results are presented in Table \ref{tab:results_few_img_to_3D}.

\begin{table}[!h]
    \centering
    \caption{\textbf{Single-shot Few Image Scene Reconstruction Comparison.} Due to occlusion artifacts we also compute all metrics for ZoeDepth and Metric3D while masking out regions occluded without points, indicated with a ‡. For a fair comparison only the unmasked values are considered for ranking the methods.}
    \vspace{-0.2cm}
    \label{tab:results_few_img_to_3D}
    \setlength{\tabcolsep}{2pt}
    \begin{tabular}{l c c c c}
        \toprule
          Methods & \textbf{PSNR} $\uparrow$ & \textbf{SSIM}$\uparrow$ & \textbf{LPIPS}$\downarrow$ & \textbf{DRMSE}$\downarrow$ \\
         \midrule
          ZoeDepth \cite{bhat2023zoedepth} & 5.466  & 0.254 &   \cellcolor{tabthird}0.563 &  11.728 \\
          ZoeDepth$^{\text{‡}}$  \cite{bhat2023zoedepth} & 14.202  &  0.661 &   0.292 &  9.378 \\
          Metric3D \cite{hu2024metric3d} & 6.314  & 0.296 &  \cellcolor{tabsecond}0.554 &  \cellcolor{tabsecond}10.049 \\
          Metric3D$^{\text{‡}}$ \cite{hu2024metric3d} & 13.699 &  0.600 & 0.336 & 8.655 \\
          NeRFacto \cite{nerfstudio_Tancik_2023}  & 10.943  &  0.298 &  0.791 & -- \\
          K-Planes \cite{fridovich-keil_k-planes_2023} & 11.356  &  0.463 &   0.633 & -- \\
          SplatFacto \cite{ye2023mathematical} & 11.607  &  0.486 &   0.658 &  -- \\
          PixelNeRF \cite{yu_pixelnerf_2021} & \cellcolor{tabthird}14.500 & \cellcolor{tabthird}0.550 & 0.652 & 19.235 \\
          SplatterImage \cite{szymanowicz23splatter}  & \cellcolor{tabsecond}17.791  & \cellcolor{tabsecond}0.580 &  0.568  & \cellcolor{tabthird}11.049 \\ 
          6Img-to-3D \cite{gieruc20246imgto3d} & \cellcolor{tabfirst}18.682 & \cellcolor{tabfirst}0.726 & \cellcolor{tabfirst}0.451 & \cellcolor{tabfirst}6.232 \\ 
        \bottomrule
    \end{tabular}
\end{table}

For K-Planes, SplatFacto, SplatFacto-big, and NeRFacto we picked five scenes to evaluate the methods on and averaged the score across those. Those methods do not make use of data-driven priors and do not profit from training on data other than the ones relevant to the evaluation scene. PixelNeRF, SplatterImage, and 6Img-to-3D are trained across the training towns.

\section{Conclusion}\label{sec:conclusion}

We present a user-friendly 3D and 4D data generator, two ego-exo view datasets, and several benchmarks. Our presented open-source data generator enables the fast and customizable creation of dynamic 3D data tailored for various tasks. The generator allows the creation of synthetic images from camera setups commonly used in NuScenes, KITTI360, and the Waymo dataset. Currently, no vision-based 4D prediction methods exist to test on our benchmarks. Our static dataset combines vehicle-mounted outward-facing ego views and inward-facing surround vehicle camera views. It is well suited for few-image-to-3D, scene reconstruction, and novel view synthesis tasks that work with outward-facing minimally overlapping cameras. Our dynamic dataset provides a large-scale multi-view dynamic urban scene dataset with diverse camera viewpoints. We hope our data generator, the datasets, and the introduced benchmarks will fertilize new research across communities, by fostering progress toward few-image-to-3D reconstruction, 3D temporal predictions, and eventually 4D predictions.
\vspace{-0.2cm}

\paragraph{Limitations.}
The synthetic data generated with CARLA is not photorealistic. Style transfer methods \cite{richter2021enhancing, keser2021content, wang2023stylediffusion}, especially recent sim-to-real methods focusing on CARLA \cite{pasios2024carla2realtoolreducingsim2real} or the planned porting of CARLA from Unreal Engine 4 to 5 could increase the quality. Within CARLA, the dynamics model used to steer the vehicles is also somewhat limited. Finally, the dataset is not general-purpose: we focus on outdoor street scenes and driving scenes and do not cover other contexts where ego--exo data would be useful.
\vspace{-0.2cm}

\paragraph{Future Work.} The presented work could be used for evaluating and performing additional tasks, such as:
\begin{itemize}[leftmargin=*,nosep]
    \item \textbf{4D prediction and reconstruction.} Predicting appearance and geometry in 3D at future time points could increase the temporal coherence of predictions. 4D prediction tasks are still in their infancy. % Concurrent to our paper a Gaussian-based dynamic 3D prediction method was published \cite{ZhaoGaussianPrediction}, a first paper in this direction that would benefit from using our dataset.
    %\item \textbf{4D reconstruction.} Extending the presented ego--exo few-image-to-3D benchmark to incorporate multi-timestep data. The multi-view novel view synthesis benchmark could be extended in a similar way to test existing spatiotemporal reconstruction methods.
    \item \textbf{LiDAR aided few-image reconstruction.} Other sensor modalities, such as LiDAR, could support few-image-to-3D reconstruction. %, particularly when using Gaussian splatting.
    \item \textbf{3rd person view prediction.} Non-ego vehicle perspective reconstruction could aid during imitation learning and allow learning from 3rd persons driving behavior.
\end{itemize}

\paragraph{Acknowledgements} The research leading to these results is partially funded by
the German Federal Ministry for Economic Affairs and Climate Action within the
project “NXT GEN AI METHODS". The authors wish to extend their sincere
gratitude to the creators of NeRFStudio, and the CARLA simulator
for generously open-sourcing their code.

%%%%%%%%% REFERENCES
{\small
\bibliographystyle{ieee_fullname}
\bibliography{egbib}
}

\clearpage
\setcounter{page}{1}
%\maketitlesupplementary

\section*{Appendix}\label{sec:appendix}

\section{Licenses} \label{sec:licenses}

Below in Table \ref{tab:licenses} the licenses of the code and assets we make use of are listed. Neo360 is listed because we use its re-implementation of PixelNeRF.

\begin{table}[ht!]
        \caption{\textbf{Licenses.}}
    \label{tab:licenses}
    \centering
    \begin{tabularx}{1\linewidth}{l c c}
        \toprule
         Item &  License \\
         \midrule
          CARLA code &   MIT   \\
          CARLA assets & CC-BY  \\
          NeRFStudio &  Apache-2.0 \\
          PixelNeRF & BSD-2-Clause  \\
          SplatterImage &  BSD-3-Clause   \\
          6Img-to-3D   &  BSD-3-Clause \\
          Neo360 & Non-commercial attribution \\
         \bottomrule
    \end{tabularx}
\end{table}

\section{Dataset Details} \label{sec:dataset-details}

\subsection{Extended Dataset Description}

The static (3D) dataset encompasses 212k inward—and outward-facing vehicle images, while our dynamic (4D) dataset contains 16.8M images from 10k trajectories, each sampled at 100 points in time with egocentric and exocentric images. Data for the static dataset is collected from 2002 scenes, and for the dynamic dataset, from 498 scenes. Because the static and the dynamic datasets differ in amount of vehicles that are equipped with sensors they differ in their composition as highlighted in Table \ref{tab:number_of_views}.

\begin{table}[ht!]
        \caption{\textbf{Number of Views.}}
    \label{tab:number_of_views}
        \centering
    \begin{tabularx}{1\linewidth}{l c c}
        \toprule
         Dataset & Egocentric & Exocentric \\
         \midrule
          Static (all) &  12k & 200k    \\
          Static (per vehicle) & 12k & 200k  \\
          Dynamic (all) & 6.3M  & 10.5M   \\
          Dynamic (per vehicle) & 300k  & 500k  \\
         \bottomrule
    \end{tabularx}
\end{table}

The uncompressed static dataset has a total size of 437 GB and took 132 hours of GPU time to be generated. Per scene, this corresponds to a size of 0.218 GB and a generation time of 4 minutes. The uncompressed dynamic dataset is 1673 GB large and took 390 hours of GPU time to be generated. Since the dataset contains images from 498 scenes and 21 vehicles per scene this results in 10458 sequences. Each sequence with a length of 100 timesteps has a size of 0.160 GB and required 2.23 minutes to generate.

\subsection{Directory Setup}

Each of the datasets (static and dynamic) is organized in the following way: towns, weather, ego vehicle type, ego-position (spawn point), timesteps, vehicles in the scene, and finally folders containing the actual sensor measurements, transforms, and camera information.

\begin{forest}
  for tree={
    font=\ttfamily,
    grow'=0,
    child anchor=west,
    parent anchor=south,
    anchor=west,
    calign=first,
    edge path={
      \noexpand\path [draw, \forestoption{edge}]
      (!u.south west) +(7.5pt,0) |- node[fill,inner sep=1.25pt] {} (.child anchor)\forestoption{edge label};
    },
    before typesetting nodes={
      if n=1
        {insert before={[,phantom]}}
        {}
    },
    fit=band,
    before computing xy={l=15pt},
  }
[Dynamic Dataset
    [Town01
        [ClearNoon
            [vehicle
                [spawn\_point\_1
                    [step\_0 
                        [370
                            [nuscenes 
                                [sensors 
                                    [0\_depth.png]
                                    [0\_instance\_seg.png]
                                    [0\_lidar.ply]
                                    [0\_optical\_flow.png]
                                    [0\_rgb.png]
                                    [0\_semantic\_seg.png]
                                ]
                                [transforms
                                    [transforms.json]
                                ]
                                [camera\_info.json]
                            ]
                            [nuscenes\_lidar]
                            [sphere]
                        ]
                        [371]
                        [372]
                    ]
                [step\_1]
                [step\_2]
            ]
            [spawn\_point\_2]
            [spawn\_point\_3]
            ]
        ]
    ]
]
\end{forest}

\section{Benchmark Details} \label{sec:benchmark-details}

\begin{figure*}[ht!]
    \centering
    \includegraphics[width=1\linewidth]{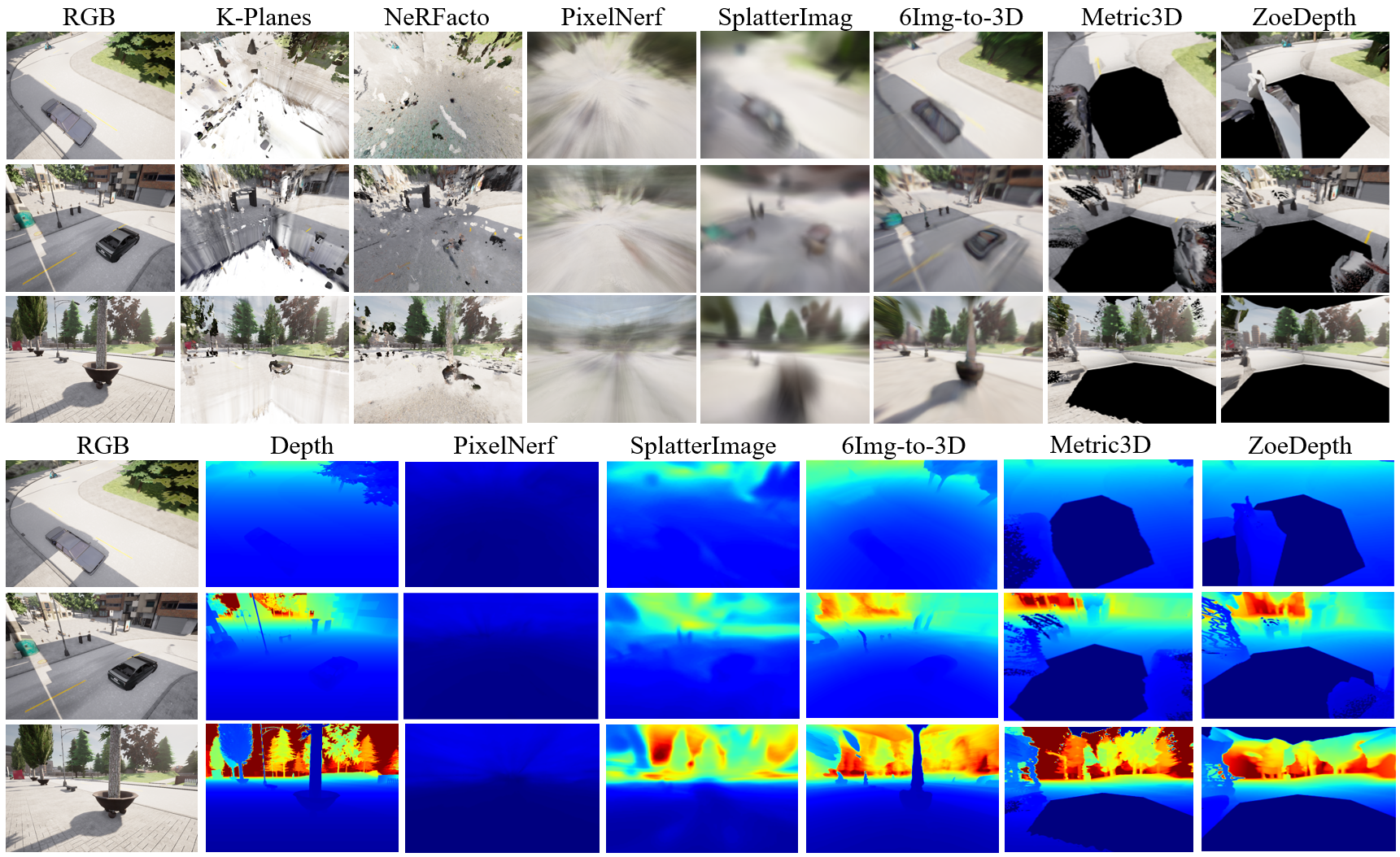}
    \caption{Qualitative Results for the single-shot few image scene reconstruction methods.}
    \label{fig:qualitative_results_overview}
\end{figure*}

\begin{figure*}
    \centering
    \includegraphics[width=1\linewidth]{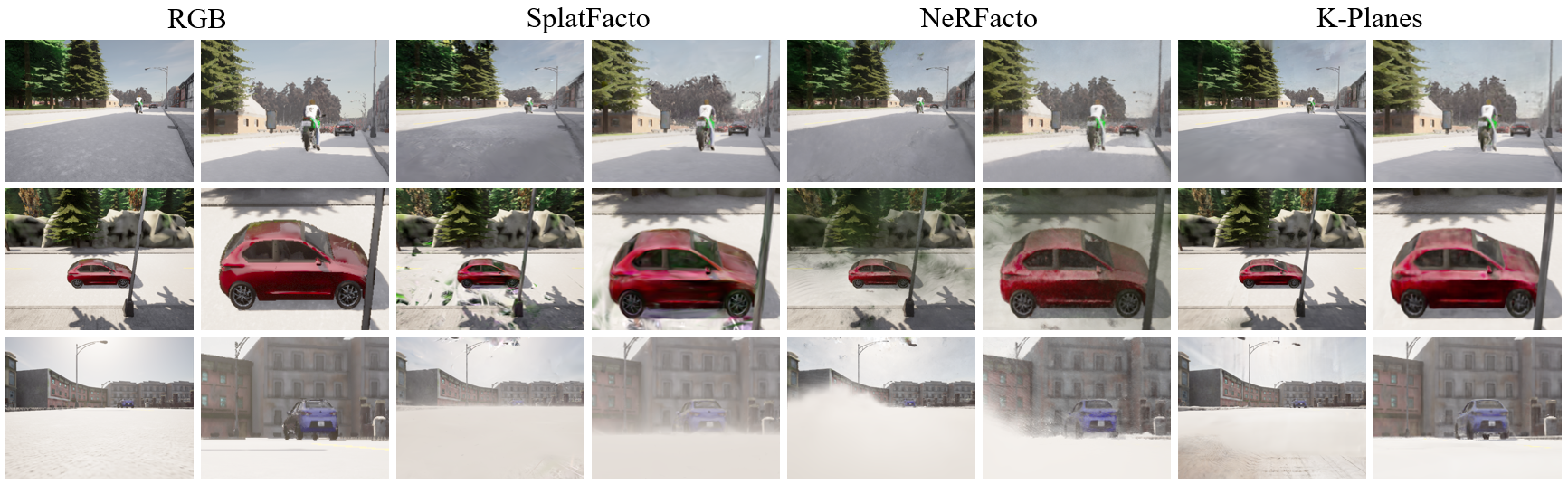}
    \caption{Qualitative Results for the multi-view per scene optimization methods. }
    \label{fig:qualitative_multiview_results}
\end{figure*}

\subsection{Qualitative Results}

\paragraph{Multi-view Novel View Synthesis.} Figure \ref{fig:qualitative_multiview_results} compares the qualitative results of Splatfacto, Nerfacto, and K-Planes. Our analysis shows that K-planes generalizes best both quantitatively and qualitatively, as demonstrated by the minimal presence of floaters. Interestingly, SplatFacto significantly outperforms both other methods on the training set but performs worst on the test set, as shown in Table \ref{tab:results_multi_view_extended}. We hypothesize that K-Planes' planar representation provides geometric regularization that enhances generalization performance.

\begin{table*}[!ht]
    \centering
    \caption{\textbf{Training and Testing result comparison of Multi-view Novel View Synthesis Methods.} }
    \vspace{-0.2cm}
    \label{tab:results_multi_view_extended}
    \setlength{\tabcolsep}{4pt}
    \begin{tabular}{l ccc ccc}
        \toprule
          & \multicolumn{3}{c}{\textbf{Train}} & \multicolumn{3}{c}{\textbf{Test}} \\
         \cmidrule(lr){2-4} \cmidrule(lr){5-7}
          Methods & \textbf{PSNR}$\uparrow$ & \textbf{SSIM}$\uparrow$ & \textbf{LPIPS}$\downarrow$ & \textbf{PSNR}$\uparrow$ & \textbf{SSIM}$\uparrow$ & \textbf{LPIPS}$\downarrow$  \\
         \midrule
         SplatFacto \cite{kerbl20233d} & \cellcolor{tabfirst}44.019 & \cellcolor{tabfirst}0.984  & \cellcolor{tabfirst}0.014 & \cellcolor{tabthird}24.458  &  \cellcolor{tabsecond}0.806 &  \cellcolor{tabfirst}0.210  \\
        NeRFacto \cite{nerfstudio_Tancik_2023} &  \cellcolor{tabsecond}36.206 &  \cellcolor{tabsecond}0.930  & \cellcolor{tabsecond}0.091 & \cellcolor{tabsecond}24.936  &  \cellcolor{tabthird}0.804 &  \cellcolor{tabsecond}0.227 \\
       K-Planes \cite{fridovich-keil_k-planes_2023} & \cellcolor{tabthird}29.827 & \cellcolor{tabthird}0.820 & \cellcolor{tabthird}0.254 &   \cellcolor{tabfirst}25.744 & \cellcolor{tabfirst}0.816  &  \cellcolor{tabthird}0.239  \\
        \bottomrule
    \end{tabular}
\end{table*}

\paragraph{Single-shot Few-Image Scene Reconstruction.} In Figure \ref{fig:qualitative_results_overview}, the methods performing single-shot few-image scene reconstruction. K-Planes, NeRFacto, and PixelNerf visibly struggle to reconstruct the scene. Where the unprojected depth maps obtained via Metric3D and ZoeDepth result in pixel values good results are obtained. The few-image SplatterImage and 6Img-to-3D perform reasonably well. Due to their low performance, we do not visualize SplatFacto K-Planes and NeRFacto at the bottom part.

\subsection{Training Details} \label{sec:training-details}

\textbf{K-Planes} We train each of the models for 30k steps on a single Tesla T4 GPU with 16GB of VRAM. We follow the model's default NeRFStudio \cite{nerfstudio_Tancik_2023} settings for training. Near and far bounds of the scene are adjusted to 0.1 to 60 to best accommodate the scenes. Additionally, scene contraction is applied. The training took around 1.5 hours per model.

\textbf{NeRFacto} We train each of the models for 30k steps on a single Tesla T4 GPU with 16GB of VRAM. We follow the model's default NeRFStudio \cite{nerfstudio_Tancik_2023} settings for training. We disable the model's use of an appearance embedding since those lead to problems during the evaluation, and we also deactivate the camera pose optimization because we already provide the model with ground truth poses. The near and far bounds are set to 0.1 and 60. Each model is trained for a total of 1 hour.

\textbf{SplatFacto} We train each of the models for 30k steps on a single Tesla T4 GPU with 16GB of VRAM. We again follow the model's default NeRFStudio \cite{nerfstudio_Tancik_2023} settings for training. The model took a total of 20 minutes to train.

\textbf{PixelNeRF} We train PixelNeRF for 100k steps on a Nvidia A40 GPU with 42GB of VRAM, with an Adam optimizer \cite{kingma2017adam} and a learning rate of $1\mathrm{e}\text{-}3$. Total training time accumulates to five days.

\textbf{SplatterImage} We train SplatterImage for a total of five days across five 3090 GPUs with 24GB of VRAM. During training the supervision images are scaled to $128 \times 128$ pixels. We use the multi-input image variant of the model to accommodate all six input views.

\textbf{6Img-to-3D} We train 6-Img-to-3D, following 
their \cite{gieruc20246imgto3d} process, with a Nvidia A40 GPU with 42GB of VRAM for 100 epochs with an Adam optimizer \cite{kingma2017adam}, a learning rate of $5\mathrm{e}\text{-}5$, and a cosine scheduler \cite{loshchilov2017sgdr} with 1000 warmup steps. Each epoch consists of 1900 steps, each comprising a new scene and three randomly sampled views as supervision, scaled to $64 \times 48$ pixels. The total training of the model is five days.

\textbf{ZoeDepth} and \textbf{Metric3D} were not fine-tuned using our data. For the single-shot few image reconstruction task, we tested both monocular depth estimation as a baseline. We obtained a depth map for each of the six ego input images resized to $842 \times 842$ to fit the model. Since camera intrinsics and extrinsic are known, we can use the depth maps to project the image pixels into space to obtain a colored point cloud (sometimes also referred to as 2.5D). The obtained colored point cloud can now be used to rasterize novel exo views.

\section{Leaderboard}

We will actively maintain a leaderboard on the project page accompanying our SEED4D paper. We welcome contributions to one of the proposed benchmarks or other submissions using the datasets. Submissions can be made by contacting the first author.

\section{Hosting, licensing, and maintenance plan}

\paragraph{Hosting.} To find the latest hosting information of our datasets please see our project page \href{https://seed4d.github.io/}{here}.

\paragraph{Licensing.} Below in Table \ref{tab:our-licenses} the licenses of the code and assets we are publishing are listed.

\begin{table}[ht!]
        \caption{\textbf{Own Licenses.}}
    \label{tab:our-licenses}
        \centering
    \begin{tabularx}{0.7\linewidth}{l c}
        \toprule
         Item &  License \\
         \midrule
          Data generator code &   BSD-3-Clause   \\
          Static dataset & CC BY-SA 4.0 \\
          Dynamic dataset &  CC BY-SA 4.0 \\
          ArXiv paper & CC BY 4.0  \\
         \bottomrule
    \end{tabularx}
\end{table}

\paragraph{Responsibility Statement} We believe that our datasets comply with existing licenses and have adhered to their terms and conditions. Despite our careful attention to these requirements, we acknowledge that any responsibility for any potential rights violations remains solely ours. We take accountability for ensuring that all content and actions are following legal and ethical standards.

\section{Data Generation Details} \label{sec:generator-details}

\subsection{Carla Towns}

The towns available within Carla vary in scenery, road structure, and size, with key characteristics highlighted below:

\textbf{Town 1}: Town 1 is a compact environment divided by a river with several small bridges. The road network includes numerous T-junctions and a variety of buildings, both residential and commercial, surrounded by coniferous trees.

\textbf{Town 2}: Town 2 consists of a mix of residential and commercial areas, including a central park, apartment buildings, a church, and a gas station. The road network is composed of T-junctions and tree-lined streets.

\textbf{Town 3}: Town 3 is an urban area featuring a central roundabout, raised metro tracks, and a diverse mix of commercial and residential buildings. The road network includes four-way junctions, T-junctions, an underpass, overpasses, and cul-de-sacs.

\textbf{Town 4}: Town 4 is a small town with a ring road in a "figure of 8" configuration that includes an underpass and overpass. The town features commercial and residential buildings, tree-lined streets, nearby snow-capped mountains, and a pedestrian shopping arcade.

\textbf{Town 5}: Town 5 is an urban setting with multilane roads and a raised highway forming a ring road. The layout includes commercial buildings, a construction site, and a large carpark, with roads passing beneath one of the buildings.

\textbf{Town 6}: Town 6 is a low-density area with wide 4-6 lane roads interconnected by slip roads and junctions, including Michigan Left configurations. The layout features designated turning lanes and cul-de-sacs.

\textbf{Town 7}: Town 7 represents a rural area with cornfields, barns, grain silos, and windmills. Its road network is simple, with unmarked roads, a small residential street, and a short bridge over a water body.

\textbf{Town 10}: Town 10 is an urban grid layout with a mix of junction types, including yellow-box intersections and dedicated turning lanes. The town features waterfront promenades, tree-lined boulevards, skyscrapers, and public buildings such as a museum.

More information about the Carla simulator can be found in the official  
\href{ https://carla.readthedocs.io/en/latest/}{Carla documentation} \cite{Carla_Dosovitskiy17}.

\subsection{Camera Poses}

The algorithm to obtain the spherical Fibonacci lattice is described in detail in Algorithm \ref{alg:points}. The procedure equally spaces points on a half-disk. The obtained points are then translated into Carla world coordinates. To obtain the proper camera orientations, we introduce the procedure presented in Algorithm \ref{alg:angles}.

\begin{algorithm}[ht!]
\caption{Exocentric camera coordinates}\label{alg:points}
% Leave the scaling parameter out here
\begin{algorithmic}[1]
    \Procedure{Create\_sphere}{$N$}       \Comment{N points}
        \State $\phi = 3\pi - \sqrt{5}$ 
        \State $ys \gets \text{linspace($0$, $1$, $N$)} $
        \State $points \gets \text{empty list}$
        \State $idx \gets 0$
        \For{\texttt{y in ys}}
            \State $x =  \cos(\phi \cdot idx) \cdot \sqrt{1 - y^2}$
            \State $y =  \sin(\phi \cdot idx) \cdot \sqrt{1 - y^2}$
            \State $z = \text{y}$
            \State $points[idx] = [x, y, z]$
            \State $idx = idx + 1$
        \EndFor
        \State \textbf{return} $points$     \Comment{dim: $N$ x $3$}
    \EndProcedure
\end{algorithmic}
\end{algorithm}

\begin{algorithm}[ht!]
\caption{Exocentric camera orientation}\label{alg:angles}
\begin{algorithmic}[1]
    \Procedure{Create\_sphere}{$points$}
        \State $pitchs, yaws \gets \text{empty lists}$
        \State $idx \gets 0$
        \For{\texttt{point in points}}
            \State $x, y, z \gets point$
            \State $pitch = \arcsin(z) $
            \State $yaw = sign(x) \cdot \arccos(\frac{y}{x^2 + y^2})^{0.5}$
            \State $pitchs[idx], yaws[idx] = pitch, yaw$
            \State $idx = idx + 1$
        \EndFor
        \State \textbf{return} $pitches, yaws$
    \EndProcedure
\end{algorithmic}
\end{algorithm}

\section{Style transfer}

\begin{figure*}[ht!]
    \centering
    \includegraphics[width=0.8\textwidth]{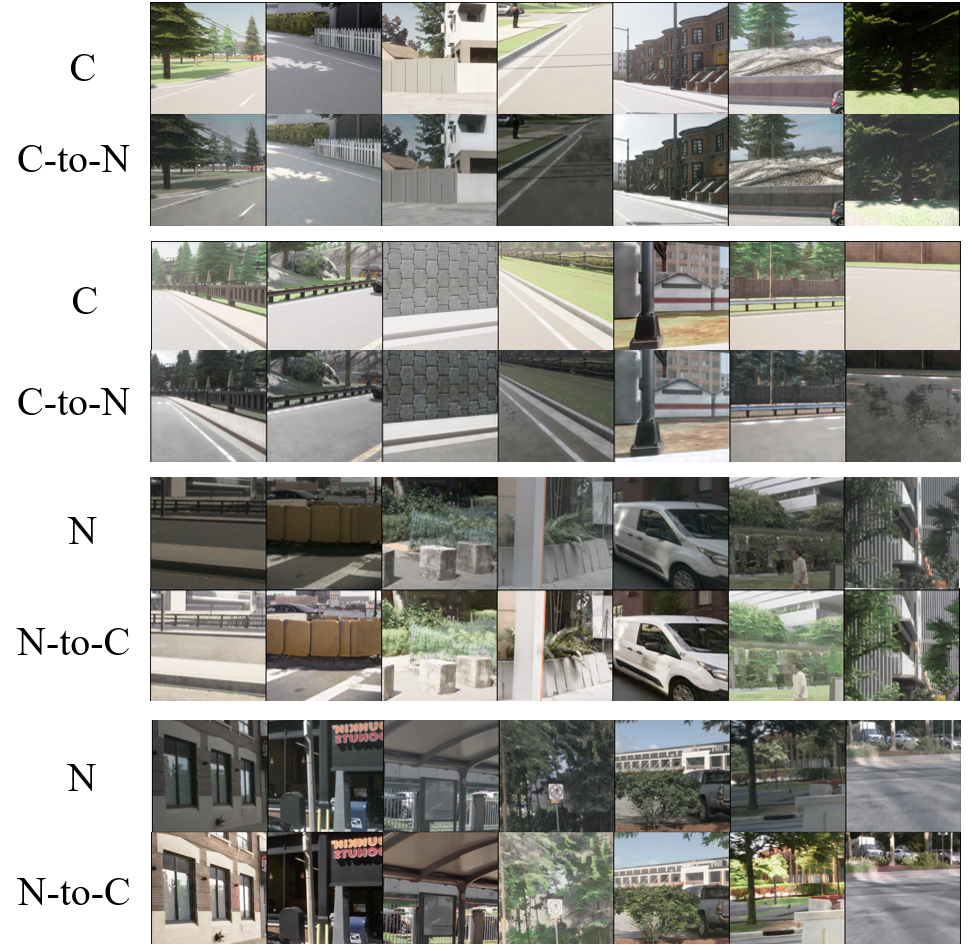}
    \caption{Example style transfer results. 'C' denotes Carla images, 'C-to-N' indicates a transfer from the Carla domain to the NuScenes domain. 'N' indicates NuScenes views, and 'N-to-C' signifies images transferred from the NuScenes domain into the Carla domain. Note: The cycle consistency step, into the original domain, is not illustrated here.}
    \label{fig:style_transfer}
\end{figure*}

We experimented with existing style transfer methods to reduce the domain gap between Carla and NuScenes' images. The results in Figure \ref{fig:style_transfer} are obtained using a CylceGan-based framework as proposed in \cite{keser2021content}. The checkpoint of our trained model will be made available.

\section{Dataset Visualization.}

\begin{figure*}
    \centering
    \includegraphics[width=0.8\linewidth]{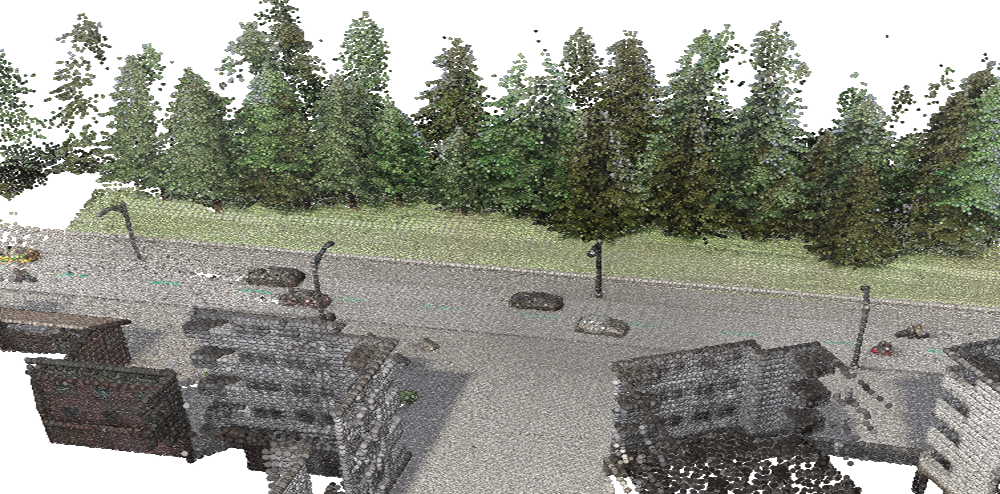}
    \caption{Colored 3D point cloud generated from RGB images, depth maps, camera intrinsics, and extrinsics.}
    \label{fig:pointcloud}
\end{figure*}

\begin{figure*}[ht!]
    \centering
    \includegraphics[width=\linewidth]{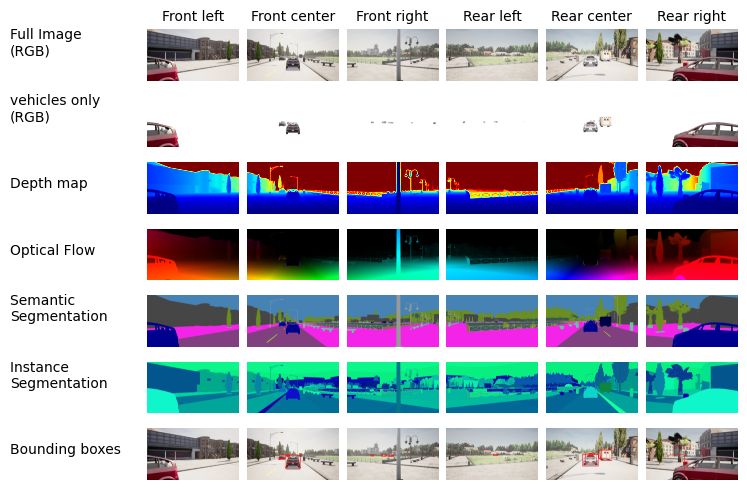}
    \caption{An overview of six egocentric cameras and their associated sensor measurements.}
    \label{fig:sensor_overview}
\end{figure*}

\begin{figure*}[ht!]
    \centering
    \includegraphics[width=0.8\textwidth]{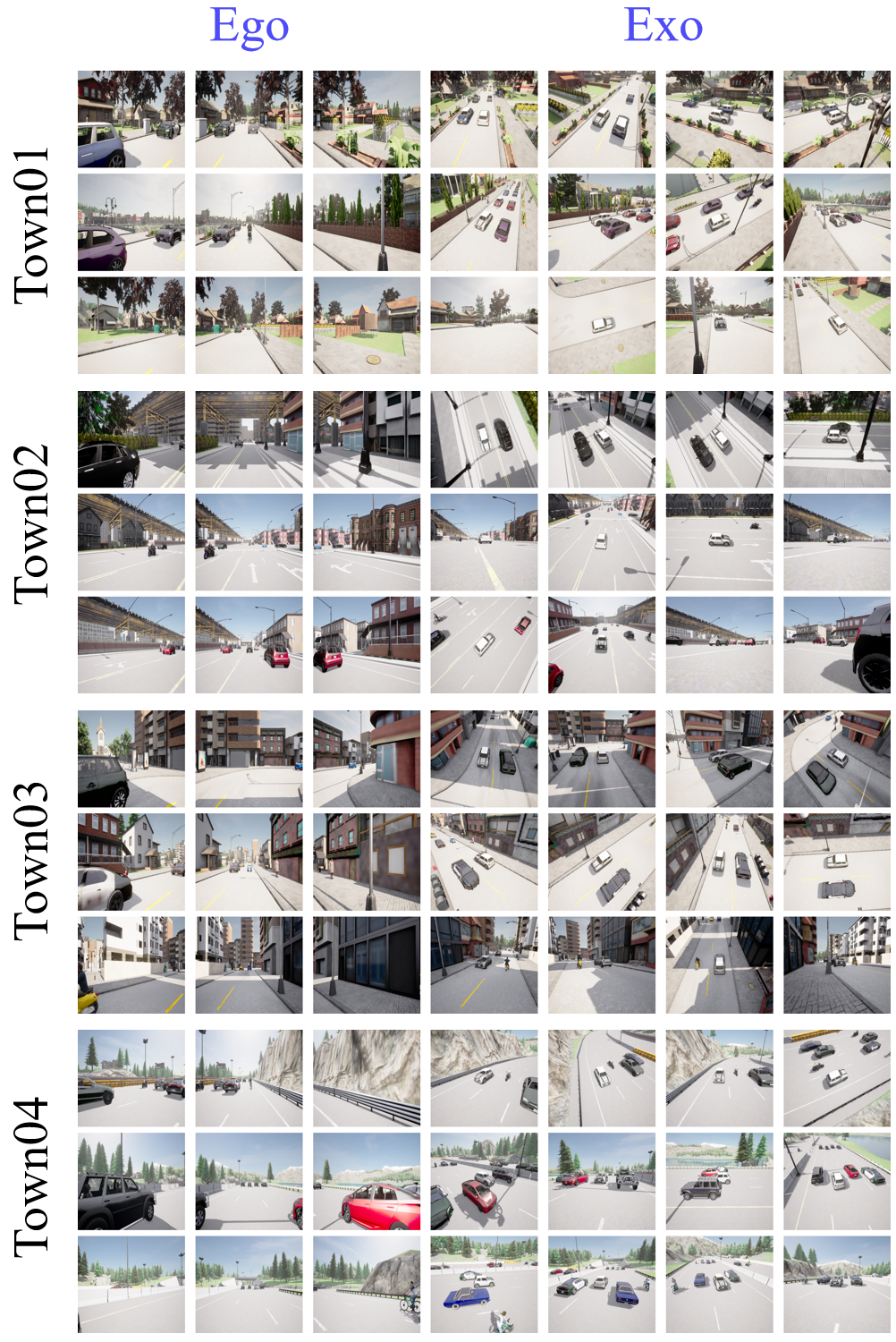}
    \caption{Samples from the Static Ego--Exo Dataset showing towns 1 to 4. The egocentric images show front left, front center, and front right views. The exocentric views are randomly sampled.}
    \label{fig:town01-town04-static}
\end{figure*}

\begin{figure*}[ht!]
    \centering
    \includegraphics[width=0.8\textwidth]{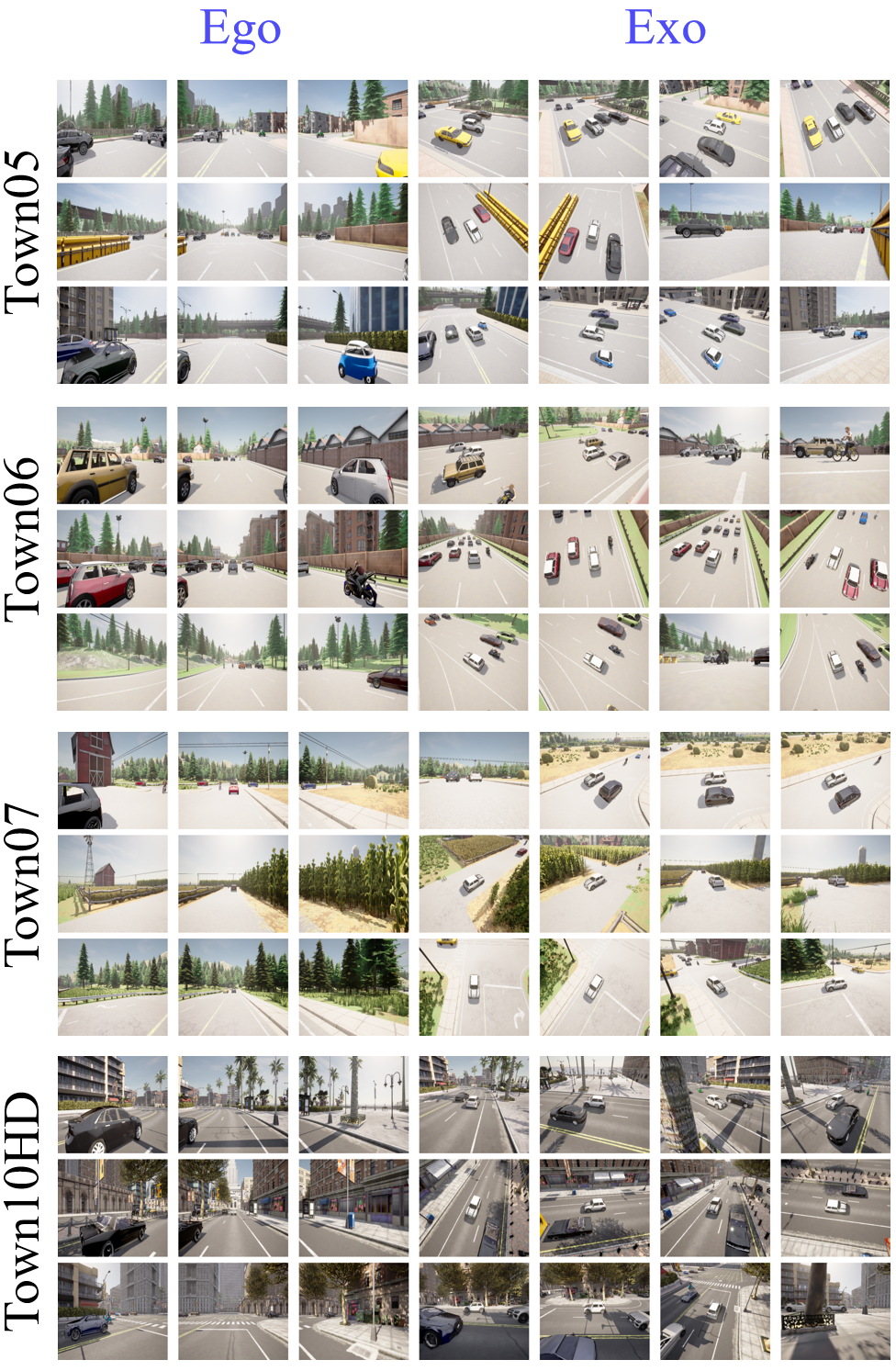}
    \caption{Samples from the Static Ego--Exo Dataset showing towns 5 to 7 and 10HD.  The egocentric images show front left, front center, and front right views. The exocentric views are randomly sampled.}
    \label{fig:town05-town10HD-static}
\end{figure*}

\begin{figure*}[ht!]
    \centering
    \includegraphics[width=0.8\textwidth]{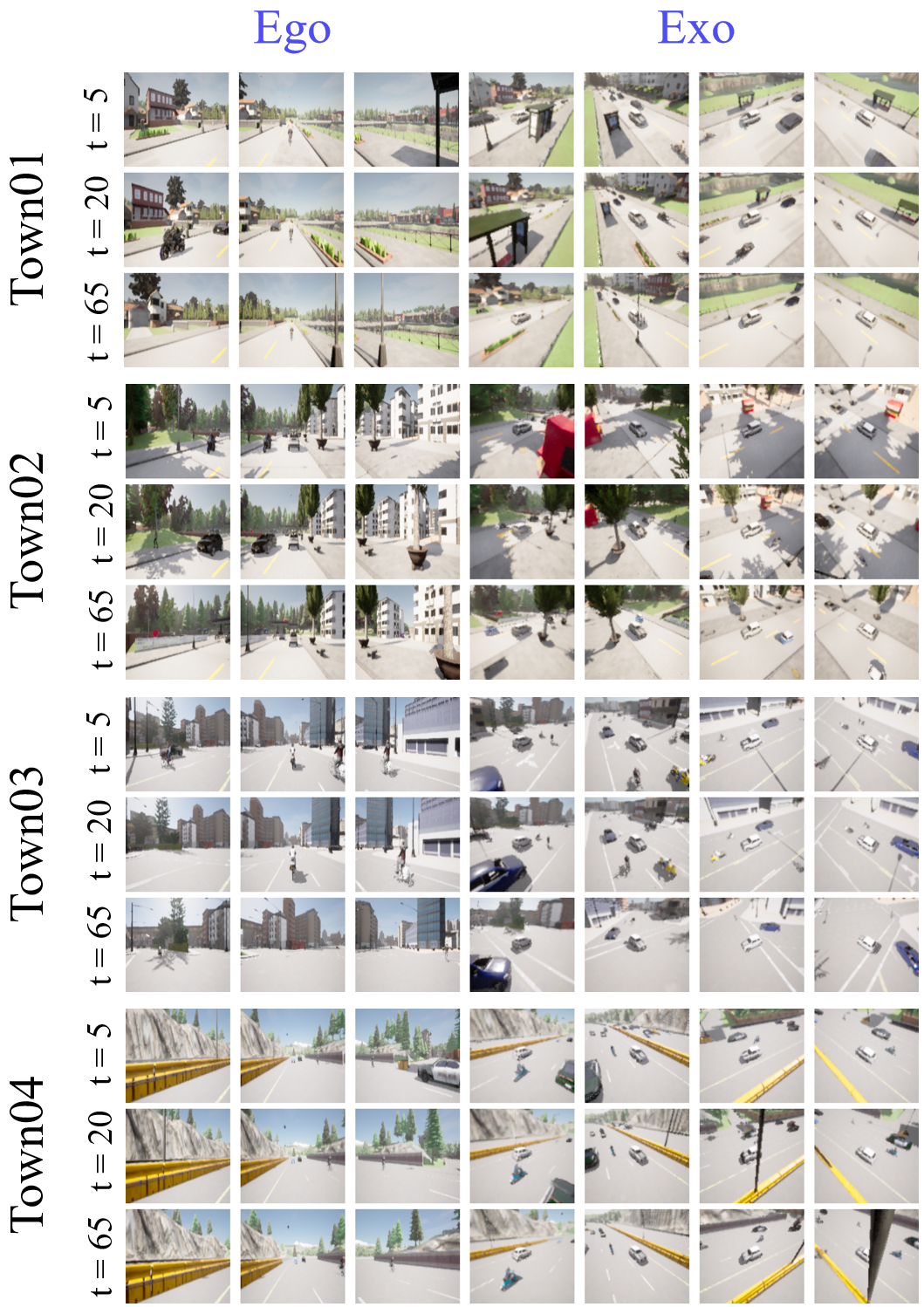}
    \caption{Samples from the Dynamic Ego--Exo Dataset showing towns 1 to 4 for timepoints 5, 20, and 65. The egocentric images show front left, front center, and front right views. The four exocentric views have the same relative pose across all samples.}
    \label{fig:town01-town04-dyn}
\end{figure*}

\begin{figure*}[ht!]
    \centering
    \includegraphics[width=0.8\textwidth]{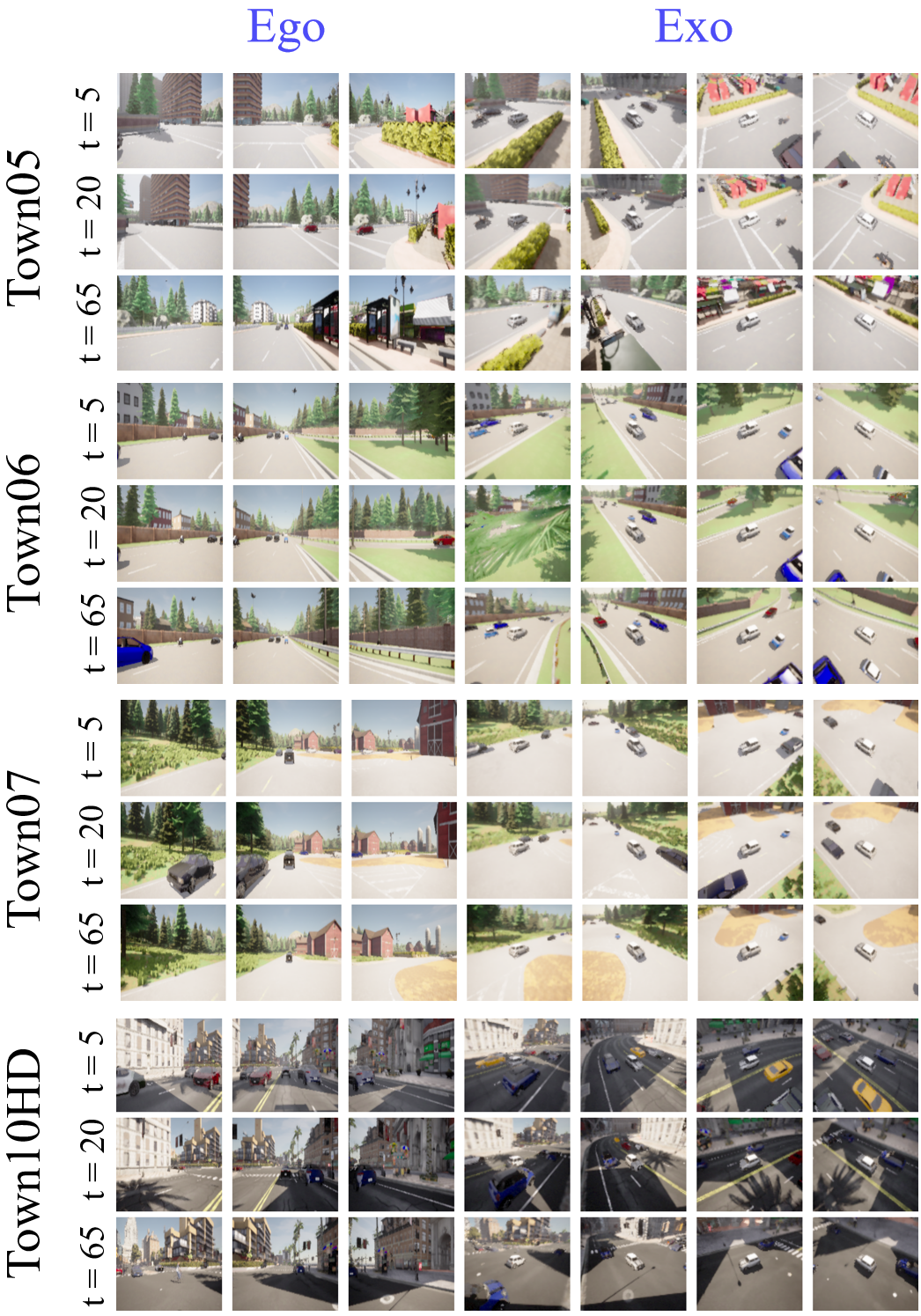}
    \caption{Samples from the Dynamic Ego--Exo Dataset showing towns 1 to 4 for timepoints 5, 20, and 65. The egocentric images show front left, front center, and front right views. The four exocentric views have the same relative pose across all samples.}
    \label{fig:town05-town10HD-dyn}
\end{figure*}

All full RGB images are paired with depth maps, optical flow, segmentation maps, and instance segmentation images. Since all values are ground truth, they can, for example, be used to generate a colored 3D point cloud using the camera's extrinsics and intrinsics, as shown in Figure \ref{fig:pointcloud}. The sensory setup for an egocentric view is visualized in \ref{fig:sensor_overview}. Figure \ref{fig:town01-town04-static} and Figure \ref{fig:town05-town10HD-static} the static ego--exo dataset is visualized. Figure \ref{fig:town01-town04-dyn} and Figure \ref{fig:town05-town10HD-dyn} display the dynamic ego--exo dataset.

\end{document}